\documentclass[sigplan, nonacm, 10pt]{acmart}

\usepackage[framemethod=TikZ]{mdframed}

\usepackage[english]{babel}
\usepackage{blindtext}
\usepackage{algorithmic}
\usepackage{graphicx}
\usepackage{textcomp}
\usepackage{xcolor}
\usepackage{wrapfig}    
\usepackage{amsfonts}
\usepackage{booktabs}
\usepackage{siunitx}
\usepackage{tablefootnote}
\usepackage{pgfplotstable} 
\usepackage{multirow}
\usepackage{paralist} % for inline list
\usepackage{amsmath}
\usepackage{enumitem}
\usepackage{caption} 
\usepackage{float}
\usepackage{subcaption}
\usepackage{hyperref}
\usepackage{enumerate}
\usepackage{enumitem}
\usepackage{layouts}
\usepackage{relsize}
\usepackage{dblfloatfix}

\usepackage{color, colortbl}
\usepackage{balance}

\AtBeginDocument{%
  \providecommand\BibTeX{{%
    \normalfont B\kern-0.5em{\scshape i\kern-0.25em b}\kern-0.8em\TeX}}}

\setcopyright{acmlicensed}
\copyrightyear{2018}
\acmYear{2018}
\acmDOI{XXXXXXX.XXXXXXX}

\acmConference[RADIO '24]{Make sure to enter the correct
  conference title from your rights confirmation emai}{June 03--05,
  2018}{Woodstock, NY}
\begin{document}

\title{PickLLM: Context-Aware RL-Assisted Large Language Model Routing}

\author{Dimitrios Sikeridis, Dennis Ramdass, Pranay Pareek}
\email{{dimitrios.sikeridis,dennis.ramdass,pranay.pareek}@broadcom.com}
\affiliation{%
  \institution{AI and Advanced Services, VMware Cloud Foundation (VCF) Division, Broadcom}
  \city{Palo Alto}
  \state{California}
  \country{USA}
}

\begin{abstract}
Recently, the number of off-the-shelf Large Language Models (LLMs) has exploded with many open-source options. This creates a diverse landscape regarding both serving options (e.g., inference on local hardware vs remote LLM APIs) and model heterogeneous expertise. However, it is hard for the user to efficiently optimize considering operational cost (pricing structures, expensive LLMs-as-a-service for large querying volumes), efficiency, or even per-case specific measures such as response accuracy, bias, or toxicity. Also, existing LLM routing solutions focus mainly on cost reduction, with response accuracy optimizations relying on non-generalizable supervised training, and ensemble approaches necessitating output computation for every considered LLM candidate. In this work, we tackle the challenge of selecting the optimal LLM from a model pool for specific queries with customizable objectives. We propose PickLLM, a lightweight framework that relies on Reinforcement Learning (RL) to route on-the-fly queries to available models. We introduce a weighted reward function that considers per-query cost, inference latency, and model response accuracy by a customizable scoring function. Regarding the learning algorithms, we explore two alternatives: PickLLM router acting as a learning automaton that utilizes gradient ascent to select a specific LLM, or utilizing stateless Q-learning to explore the set of LLMs and perform selection with a $\epsilon$-greedy approach. The algorithm converges to a single LLM for the remaining session queries. To evaluate, we utilize a pool of four LLMs and benchmark prompt-response datasets with different contexts. A separate scoring function is assessing response accuracy during the experiment. We demonstrate the speed of convergence for different learning rates and improvement in hard metrics such as cost per querying session and overall response latency.
\end{abstract}

\maketitle

%%%%%%%%%%%%%%%%%%%%%%%%%%%%%%%%%%%%%%%%%%%%%%%%
%% 1. Introduction
%%%%%%%%%%%%%%%%%%%%%%%%%%%%%%%%%%%%%%%%%%%%%%%%

\section{Introduction}

Recently, generative Large Language Models (LLMs) have emerged as a clear trend for solving a diverse set of problems such as assistive dialogue, summarization, text classification and coding aid~\cite{chang2023survey, srivastava2022beyond}. In addition, they are becoming more and more capable in terms of understanding complex contexts and providing factual accuracy due to larger training datasets, and an increase in model sizes~\cite{wei2022emergent}. Thus, their applicability for problem-solving across diverse applications and industry domains has led to the explosion of their number with new LLMs being released daily by academia and industry alike\footnote{e.g., Hugging Face currently hosts an impressive number of models related to text generation}, both in open-source fashion (e.g., Meta's Llama~\cite{touvron2023llama}) but also as fully supported LLMs-as-a-service(e.g., OpenAI's GPT models~\cite{achiam2023gpt}).

While the observed influx of proprietary and open-source LLMs adds diversity and a plethora of options for practitioners and newly flocked users, there are no obvious selection criteria for picking between the different LLM offerings~\cite{li2024more}. Indeed, open-source LLMs tend to exhibit diverse weaknesses, strengths, and heterogeneous expertise in various domains mainly due to variations in architecture, training data input, and parameter tuning~\cite{ziegler2019fine, chang2023survey}. Still, interested users are left with a couple of options to adjust LLMs to their use cases including fine-tuning, and even training their LLM from scratch. The latter can be an expensive process in terms of computation and large volume data gathering, and even challenging to even perform due to the scarcity of GPUs in the market. Fine-tuning on the other hand is resource friendlier, but requires a specific level of expertise among engineering teams. In addition, modern LLM offerings often restrict any direct model optimization through weight adjustments, providing just access to black-box APIs. 

For all the reasons above, more and more practitioners are now relying on off-the-shelf LLMs for building their applications. However, relevant concerns are surfacing regarding the practical aspects of running them, namely rising costs, availability, inference latency, and average accuracy depending on the use case~\cite{madaan2023automix, chen2023frugalgpt, zhang2023ecoassistant}. Indeed, the latest state-of-the-art models comprise billions of parameters leading to excessive computing power needs and even environmental impact, especially in cases where the application demands high-throughput delivery. A case in point is OpenAI's ChatGPT where the same query (4K-token context) is $\approx$20 times cheaper when GPT-3.5 is used in comparison to the newer GPT-4 model (8K-token context)\footnote{\url{https://openai.com/pricing}}. Interestingly, the estimated accumulated cost of running a small business's customer support service in GPT4 can exceed \$21K per month~\cite{chen2023frugalgpt}. 

While such cutting-edge models can indeed handle the most demanding text generation tasks, the reality is that many use cases could be handled more than adequately by less-capable and thus less-expensive models. In addition, depending on the use case, the use of local open-source models or even quantized models running on CPUs can eliminate network latency and availability concerns where accuracy degradation can be tolerated. Given the observed variations in heterogeneous cost and quality, the efficient capitalization and real-time evaluation of all the available LLM solutions is quite understudied. There is, thus, the opportunity to identify the optimal LLM for a specific task (i.e., family of queries) and within a desired context (e.g., availability, latency, cost) in a lightweight, real-time manner that can also be adaptable to changes in query themes on-the-fly.

\noindent
{\bf Contributions.} In this paper, we tackle the challenge of selecting the optimal LLM from a model pool for specific queries with customizable objectives. 
Our contributions are summarized as follows:
\begin{itemize} %[leftmargin=*]
    \item We propose PickLLM, a lightweight framework that relies on Reinforcement Learning (RL) to route on-the-fly queries to available models. 
    \item We introduce a weighted reward function that considers per-query cost, inference latency, and model response accuracy by a scoring function. 
    \item We explore two alternatives: PickLLM router acting as a learning automaton that utilizes gradient ascent to select a specific LLM, or PickLLM utilizing stateless Q-learning to explore the set of LLMs and perform selection with a $\epsilon$-greedy approach.
    \item We demonstrate the speed of convergence for different learning rates, and the effect of variations on the reward weights that result in selecting the optimal LLM in terms of cost, and latency with no accuracy degradation.
\end{itemize}

\noindent
{\bf Structure.} Section~\ref{section:related} reviews related work on LLM selectors and routing. Section~\ref{section:arch} analyses the PickLLM framework, and the underlying RL mechanisms. Section~\ref{section:eval} describes our experimental setup, and presents our findings. Finally, Section~\ref{section:future} discussed future directions, while Section~\ref{section:conclusion} concludes this paper.

\section{Related Work}\label{section:related}

Existing solutions attempt optimizing LLM selection based on either strict monetary cost, or answer accuracy, rarely for both, and certainly without taking into account other throughput performance metrics. In~\cite{chen2023frugalgpt}, the authors propose the sequential use of different increasingly expensive LLMs in a cascade manner until the output of a dedicated scoring model (specifically DistilBERT) is acceptable. Similarly, AutoMix~\cite{madaan2023automix} utilizes small-size LLMs to produce an approximate correctness metric for outputs before strategically routing the queries further to larger LLMs. LLM cascades are also used in~\cite{yue2023large}, where the authors utilize the “answer consistency”~\cite{wang2022self} of the weaker LLM as the signal of the query difficulty that will drive the decision to route towards a more capable, and more expensive LLM. The requirement of the above solutions falls on potentially utilizing multiple models per input before yielding the optimal result.

Another family of solutions relies only on computationally heavy pre-training of supervised reward models or uses generic datasets for the supervised training which does not provide generalization for different user cases. For instance, the authors in~\cite{vsakota2023fly} utilize a regression model that is pre-trained on pairs of language model queries and related output scores. The work in~\cite{shnitzer2023large} utilizes existing benchmark datasets to train an LLM router model through a collection of binary classification tasks. In \cite{lu2023routing} the authors deploy reward model ranking on a query-answer training set to obtain an estimation of expertise between open-source LLMs. The normalized rewards from the training phase are subsequently used to train a routing function that will make the final routing decision of a new query in the online phase.

Finally, prior work on LLM selection relies on the selection of the LLM that generates the optimal output for the given input after producing all the possible outcomes. The authors in \cite{liu2021simcls}, and \cite{ravaut2022summareranker} propose the training of specific ranking and scoring models after considering outputs from every examined LLM. The work in \cite{jiang2023llm} introduces a pairwise comparison to evaluate differences between candidate LLM outputs from the same query. Following that, a fusing mechanism merges the candidate responses that ranked higher to create the final improved output. However, ranking between answer pairs, although effective, becomes infeasible for real-time applications as the LLM option space expands as each candidate has to be compared with all the rest.

Recently, a promising solution to the LLM routing problem has been proposed in~\cite{ong2024routellm}. The framework involves learning to dynamically route between a stronger model and a weaker model, thereby minimizing costs while achieving a specific performance target. The authors utilize human preference data and data augmentation techniques to enhance performance, and demonstrate significant cost savings without compromising response quality. This work highlights the potential of LLM routing to provide a cost-effective yet high-performance solution for deploying LLMs.

\begin{figure*}[t]
\centering
\includegraphics[width=0.8\textwidth]{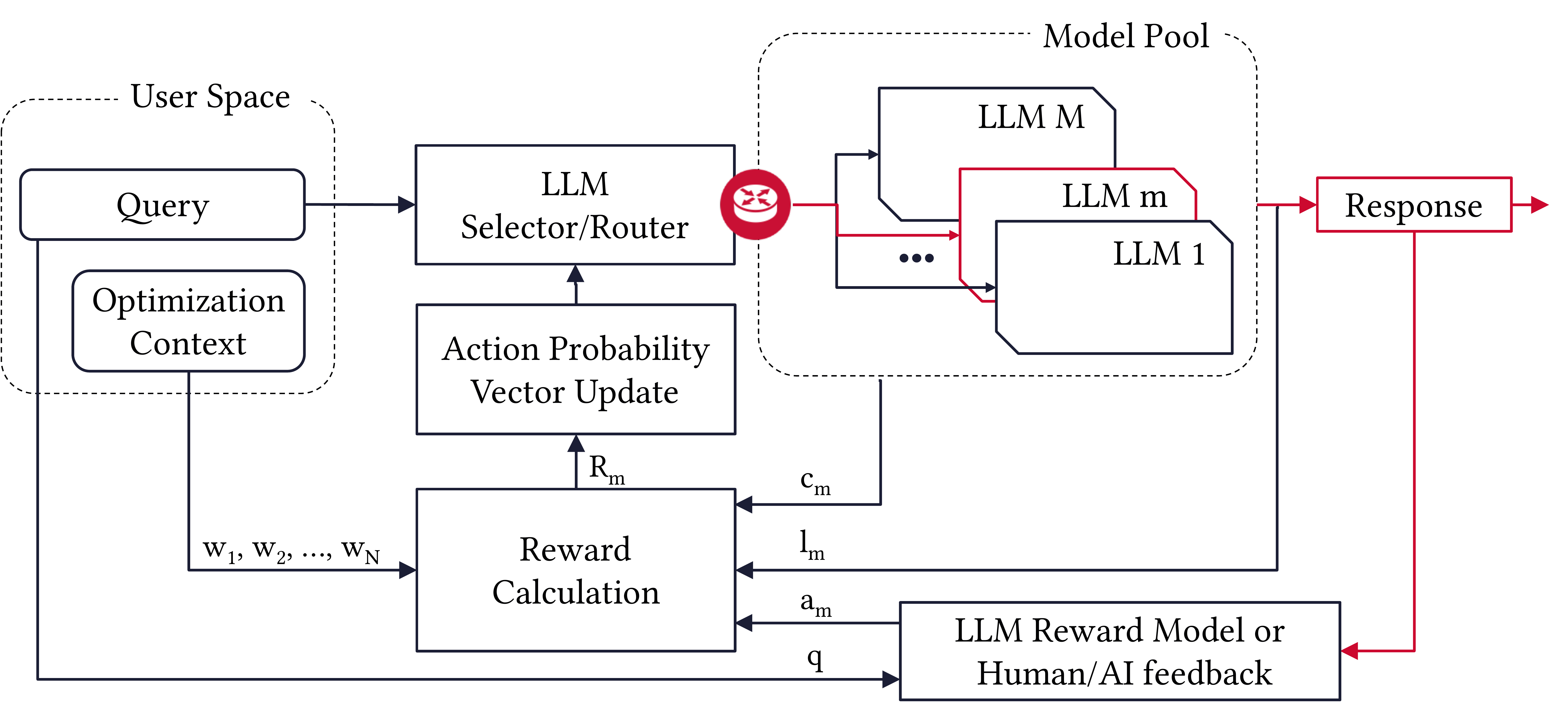}
\caption{\small SLA-based PickLLM framework.} 
\label{fig:mechanism}
\end{figure*}

\section{PickLLM}\label{section:arch}

%%%%%%%%%%%%%%%%%%%%%%%%%%%%%%%%%%%%%%
%% . Experimental Evaluation - short
%%%%%%%%%%%%%%%%%%%%%%%%%%%%%%%%%%%%%%

\subsection{Overall Framework}

In this section, we describe the basic mechanisms behind PickLLM, which specifically addresses the problem of selecting the optimal LLM for a single use case by taking into account contextual optimization choices such as running cost, or response latency. We consider a pool of $|M|$ available LLMs (set $M=\{1,...,n,...,|M|\}$) ready to receive user queries. The LLM set can be diverse, consisting of a mix of local quantized models, local models running on GPUs, or off-the-self LLMs served in cloud platforms. For each round, PickLLM routes the user query $q$ to one of the available models, observes the outcome of each response about multiple metrics, and updates the underlying LLM selection probabilities depending on the exact learning algorithms used. For each user session, i.e., a set of contextually related queries, PickLLM will converge to the most appropriate model given the optimization choices. Fig. \ref{fig:mechanism} presents an overview of the proposed framework.

\subsection{Reward Function Formulation}
Regarding the reward function, we considered a small set of hard performance-related parameters that are critical for the user's experience and perceived satisfaction. For each model $m$, we define, (a) the cost per query $c_m$, associated with the cost of running the model either locally or on the cloud, (b) the inference latency $l_m$, namely the observed inference runtime (can also account for each model's service options, variations in utilized hardware, software optimizations,  and requests spending times in queues), and (c) a response accuracy metric $a_m$, defined as the average correctness of the LLM's answer based either on human feedback or on a mixture of scoring functions. Given that, he reward function $R_m$ is defined as follows:
\begin{equation}\label{eq:reward}
R_m(a_m, c_m, l_m) = \frac{w_{a} \cdot a_m - w_{c} \cdot c_m}{w_{l} \cdot \frac{\log_{10}(l_m)}{t_{scaling}}}
\end{equation}
where $w_{a}$, $w_{c}$, and $w_{l}$ are the weights assigned to accuracy, cost, and latency, and $t_{scaling}$ is a constant used to scale latency depending on the unit of time used. Evidently, the reward reflects the "competitiveness" of each LLM $n$ as formed by response performance and user preferences. Note, that the proposed reward function is easily extensible to accommodate other optimization targets e.g., bias, toxicity, environmental impact, and misinformation, among others. HELM~\cite{liang2022holistic} offers an exhaustive list or possible LLM evaluation metrics.

\subsection{Gradient Ascent Learning}

First, we examine a gradient ascent learning approach where PickLLM acts as a stochastic learning automaton (SLA) the learns the environment by performing updates of the perceived reward~\cite{narendra2012learning}. PickLLM maintains an action probability vector $\mathbf{P}^{[i]} = [P_{1}^{[i]}, \ldots, P_{m}^{[i]}, \ldots, P_{|M|}^{[i]}]$, where $P_{m}^{[i]}$ represents the probability of PickLLM router selecting LLM $m$ for handling the upcoming query at iteration $i$. The adjustment of action probabilities follows the Linear Reward-Inaction (LRI) algorithm: when the current selection strategy is $\sigma=m$, the probability of continuing using the same LLM $m$ is updated as:
\begin{equation}
P_{m}^{[i]} = P_{m}^{[i-1]} + \beta \cdot R_{m}^{[i-1]} \cdot (1 - P_{m}^{[i-1]})
\end{equation}
while the probabilities of each one of the rest LLMs $m'$ is:
\begin{equation}
P_{m'}^{[i]} = P_{m'}^{[i-1]} - \beta \cdot R_{m}^{[i-1]} \cdot P_{m'}^{[i-1]}, \forall~m'\in M, m'\neq m 
\end{equation}
where $\beta \in (0, 1]$ denotes the learning rate, and controls the convergence of the process. PickLLM naturally converges to a single LLM for the session of queries when the action probabilities vector delta is smaller a predefined threshold $\epsilon$: i.e.,  $\max_{\sigma} |\Delta P_{m}^{[i]}| < \epsilon$
for all strategies/choices \(m\), where \(\epsilon\) is a small positive value indicating the threshold for negligible updates, ensuring that the decision-making process has stabilized. Fig. \ref{fig:mechanism} presents an overview of the proposed framework.

\begin{figure*}[t]
\centering
\includegraphics[width=0.7\textwidth]{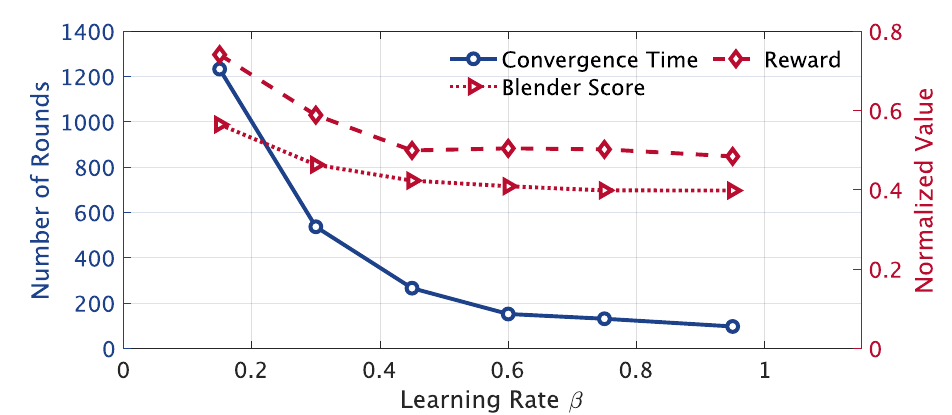}
\caption{Average of convergence time, reward and LLM-Blender score vs learning rate $\beta$ } 
\label{fig:param_b}
\end{figure*}

\subsection{Stateless $\epsilon$-greedy Q-Learning}

In addition, we study stateless $\epsilon$-greedy Q-Learning approach as an alternative mechanism for PickLLM~\cite{singh2000convergence, wilhelmi2017implications, guo2024reinforcement}. Under this scenario PickLLM maintains an action-value vector $\mathbf{Q}^{[i]} = [Q^{[i]}_{1}, \ldots, Q^{[i]}_{m}, \ldots, Q^{[i]}_{|M|}]$, where $Q^{[i]}_{m}$ approximates the utility of selecting LLM $m$ up to iteration $i$, symbolizing the expected reward \(R^{[i]}_m\) given the selection of LLM $m$, namely strategy $\sigma = m$: 
$Q^{[i]}_{m} \approx E[R^{[i]}_m | \sigma=m | i]$. The adaptation of action values follows the Q-Learning update rule:
\begin{equation}
Q^{[i]}_{m} = Q^{[i-1]}_{m} + \theta \cdot \left(R^{[i-1]}_m - Q^{[i-1]}_{m}\right)
\end{equation}
where \(\theta \in (0, 1]\) is the learning rate, influencing the magnitude of updates. For selecting LLMs, we employ an $\epsilon$-greedy approach, where PickLLM chooses the strategy $\sigma$ that maximizes the expected reward as follows:
\begin{equation*}\
    \sigma^{[i]} =  
    \begin{cases} 
        \underset{m \in M}{\arg\max}~\mathbf{Q}^{[i-1]}, & \text{with probability } 1 - \epsilon \\
    \text{random LLM $m$ from } M, & \text{with probability } \epsilon
    \end{cases}
\end{equation*}
allowing PickLLM to explore non-maximal strategies with a probability \(\epsilon\), thereby balancing exploration and exploitation. 

\begin{table*}[!t]
 \caption{PickLLM vs alternatives: GPT3 and normalized LMM-Blender scores across different dataset topics}
\renewcommand{\arraystretch}{1}
 \centering
    \begin{tabular}{l c c c c c c c c}
    \toprule
    Sub Dataset from   & \multicolumn{2}{c}{PickLLM SLA} & \multicolumn{2}{c}{PickLLM QL $\epsilon=0.1$} & \multicolumn{2}{c}{Mixtral} & \multicolumn{2}{c}{LLama2-70B} \\\cmidrule(l){2-3} \cmidrule(l){4-5} \cmidrule(l){6-7} \cmidrule(l){8-9}  HC3~\cite{guo2023close} 
     & GPT3 & LLM-Blender  & GPT3  & LLM-Blender & GPT3  & LLM-Blender &GPT3  & LLM-Blender \\
	\toprule
	 reddit\_eli5~\cite{fan2019eli5}  & 0.79  & 0.61 & 0.80 & 0.58 & \bf{0.88} & \bf{0.69} & 0.87 & 0.59 \\ 
	 open\_qa~\cite{yang2015wikiqa}   & 0.95 & 0.51 & 0.93  & \bf{0.62} & 0.95 & 0.50 & \bf{0.97} & 0.49 \\ 
	 wiki\_csai~\cite{guo2023close}    & \bf{0.91} & 0.60 &  0.93  & \bf{0.64} & 0.90 & 0.63  & 0.90 & 0.56 \\ 
	 medical~\cite{he2020meddialog}  & \bf{0.92} & \bf{0.52} & 0.90 & 0.49 & 0.89 & 0.43 & 0.91 & 0.51 \\ 
  finance~\cite{maia201818}  & 0.85 & 0.46 & 0.88 & 0.44 & 0.88 & \bf{0.48} & \bf{0.92} & 0.43 \\ 
	 \bottomrule 
	\end{tabular}
	\label{tab:mix}
\end{table*}
 
\section{Performance Evaluation}\label{section:eval}
% In this section, we describe our experimental setup, the component choices, and discuss PickLLM's performance.

\subsection{Experimental Setup}
 The experimental testbed consisted of a local host and remote execution for the LLMs utilizing an internal in-house LLM API Service. The local client running PickLLM was equipped with an Intel i9 with eight cores at 2.4 GHz each, and 32 GB of RAM. 

We used four models to construct our set $|M|$, and specifically Mistral-7B{\footnote{\tiny\url{https://huggingface.co/mistralai/Mistral-7B-Instruct-v0.1}}}~\cite{jiang2023mistral}, WizardLM-70B\footnote{\tiny\url{https://huggingface.co/WizardLM/WizardLM-70B-V1.0}}~\cite{xu2023wizardlm}, and two versions of Llama-2~\cite{touvron2023llama}, one with 13 Billion parameters\footnote{\tiny\url{https://huggingface.co/meta-llama/Llama-2-13b-chat-hf}}, and one with 70 Billion\footnote{\tiny\url{https://huggingface.co/meta-llama/Llama-2-70b-chat-hf}}. For our experiments' dataset we utilize the HC3 dataset~\cite{guo2023close} that consists of a mix of questions of varying topics and their corresponding human answers. HC3 is a collection of different individual datasets, with domains that include open-questions~\cite{fan2019eli5, yang2015wikiqa}, finance~\cite{maia201818}, and medicine~\cite{he2020meddialog}.

To simulate cost-related impact, we assume that the use of each model is associated with a normalized cost per query $c_m$, and define a cost vector $\mathbf{C_{exp}}=[0.4, 0.8, 0.7, 0.3]$ corresponding to Mistral-7B, WizardLM-70B, Llama2-70B, Llama2-13B, chosen to simulate their inference cost and diversify the selection space. As our latency metric $l_m$ we define the duration of an end-to-end communication between a user making a query and getting the response, measured in milliseconds. Finally, regarding response quality/accuracy metric, we utilize two alternatives during our experiments, namely (a) the Open Assistant's~\cite{kopf2024openassistant} deberta-v3-large-v2 reward model \footnote{\tiny\url{https://huggingface.co/OpenAssistant/reward-model-deberta-v3-large-v2}}, and (b) OpenAI's gpt-3.5-turbo\footnote{\tiny\url{https://openai.com/blog/gpt-3-5-turbo-fine-tuning-and-api-updates}} acting as a response quality judge~\cite{zheng2024judging} see specific system prompt in Appendix~\ref{appendix:a}. Both of them are normalized to yield a score  $a_m\in(0,1]$. Since the datasets employed in our experiments include human-generated answers for each query, we also utilize LLM-Blender's~\cite{jiang2023llm} pairwise reward model (PairRM)\footnote{\tiny\url{https://huggingface.co/llm-blender/PairRM}} that produces a relative quality score between two response candidates given a specific input query.

 \subsection{Evaluation and Discussion}
First, we consider how PickLLM learning rates impact the action probability convergence speed and quality of learning. Our experiment utilizes the ELI5~\cite{fan2019eli5} portion of the test dataset, fixes the reward weights to $w_{a}=0.5$, $w_{c}=0.25$, $w_{l}=0.25$, and averages 10-runs for different values of the SLA learning rate $\beta$. Fig.~\ref{fig:param_b} shows the average convergence time in terms of querying rounds for increasing values of $\beta$. For larger values of $\beta$ the convergence time is lower with a reduction of $\approx82\%$ when moving from $0.3$ to $0.9$. However, for small $\beta$ values the system usually concludes with better choices.  In this context, Fig.~\ref{fig:param_b} also shows (a) the average normalized reward gathered throughout single runs and (b) the average normalized LLM-Blender score yielded from 200 queries after the convergence of PickLLM to the optimal model. It is observed that as $\beta$ increases, PickLLM converges to models that yield smaller rewards and produce responses of lower quality. Given the fact that PickLLM aims to optimize model choice in scenarios of thematically related small-batch user queries, we opt for an average learning rate that will balance convergence time and expected performance. Thus, for all the following experiments we set the SLA learning rate to $\beta=0.5$, while the equivalent stateless Q-learning parameter $\theta$ is set to $0.7$.

\begin{figure}[t]
\centering
\includegraphics[width=\columnwidth]{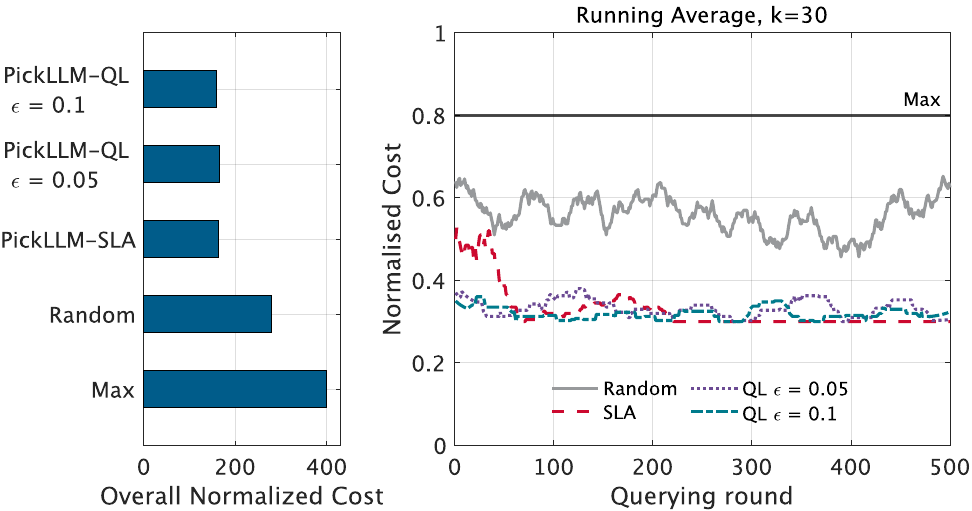}
\caption{(Right) Overall cost, (Left) Cost per query } 
\label{fig:cost}
\end{figure}

\begin{figure}[t]
\centering
\includegraphics[width=\columnwidth]{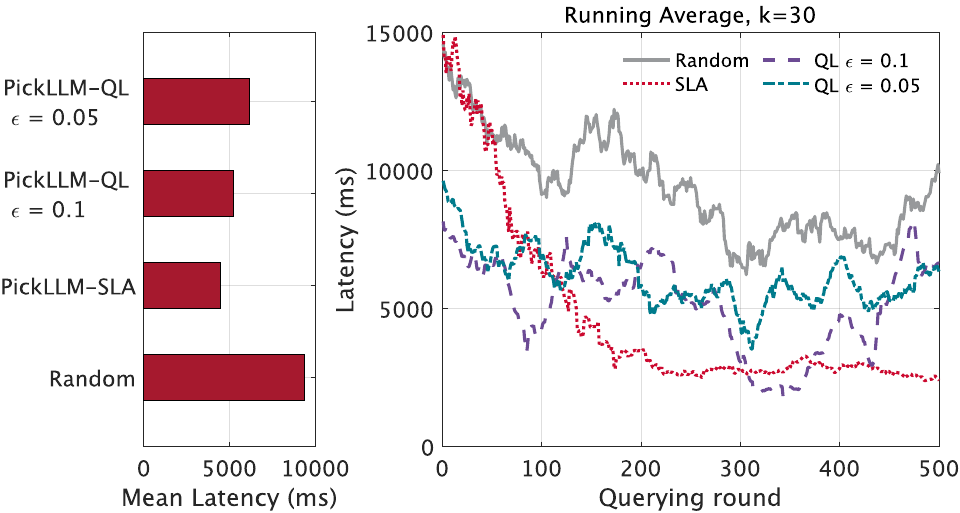}
\caption{(Right) Mean Latency (ms), (Left) End-to-end latency per query round (ms) } 
\label{fig:latency}
\end{figure}

Next, we investigate the ability of PickLLM to adjust to specific optimization contexts specifically when users aim to optimize cost for their querying sessions, or when their underlying applications are time-critical requiring optimization concerning response latency. Thus, we examine how adjusting the weights of the proposed reward function achieves this goal. First, we fix the reward weights to $w_{a}=0.2$, $w_{c}=0.6$, $w_{l}=0.2$, and run 500-query sessions from the WikiQA~\cite{yang2015wikiqa} portion of the HC3 dataset. Fig.~\ref{fig:cost}-(Right) shows the overall normalized cost of each session for different variations of PickLLM, along with two corner cases of just using the most expensive choice throughout the session, or random model selection. PickLLM leads to lower total cost, while the SLA variation reduces the cost of the session $60\%$ in comparison to the max, and less than $3\%$ compared with the use of the Q-learning variations. This is attributed to the nature of the $\epsilon$-greedy method that produces more exploration during the early portions of the session. Fig.~\ref{fig:cost}-(Left) presents a running average of the real-time cost for each query round showcasing the convergence of PickLLM to a cost-efficient final state. Results are similar when the focus is set on optimizing latency for the same query set and reward function weights of $w_{a}=0.2$, $w_{c}=0.2$, $w_{l}=0.6$. Fig.~\ref{fig:latency}-(Right) shows the average latency of each session for the same variations of PickLLM, while  Fig.~\ref{fig:latency}-(Left) presents a running average of the real-time latency for each query round. Overall, in the best case scenario, mean latency is reduced by $52\%$ over the random case, with the SLA variation showing less fluctuation over the alternatives.

Finally, we want to demonstrate the performance of PickLLM across datasets of different topics and do a comparison in terms of response performance. For the latter, two alternatives were considered: (a) using our largest model (parameter-size-wise), namely  Llama2-70B for all queries, and (b) using a mixture of experts, namely Mixtral\footnote{\tiny\url{https://huggingface.co/mistralai/Mixtral-8x7B-Instruct-v0.1}}~\cite{jiang2024mixtral} that has shown superior performance over many off-the-shelf LLMs. While the response quality of PickLLM can be as good as the underlying LLMs on the availability pool, our goal is to show that the outcome can be on par with state-of-the-art solutions while also allowing for adaption to optimization contexts set by the user. For PickLLM, we set reward function weights to $w_{a}=0.33$, $w_{c}=0.33$, $w_{l}=0.33$, and all candidates perform 500-query sessions from each one of the sub-datasets of HC3. The responses of PickLLM and both alternatives were also scored using the GPT3 judge and normalized LLM-Blender score mentioned above. Table~\ref{tab:mix} shows the final results of average GTP3 and LLM-Blender scores per session, that reflect the competitive performance of PickLLM across varying topics and contexts. The maximum scores for each sub-dataset and for each scoring method (GTP3 and LLM-Blender) are highlighted with bold.

\section{Future Work}\label{section:future}

PickLLM aims for a simple way to guide agnostic users to an LLM choice that fits their data, and operational requirements. In the next steps, we consider the move from a model-free RL mechanism to a model-based RL approach that constructs a model of the environment to predict future states and rewards. For instance, model-based contextual multi-armed bandits, and statistical learners, such as neural networks, can be used to learn and generalize value functions to predict total returns (e.g., outcome from LLM choice) given a state and action pair. This will allow us to consider user context/conversation history towards steering users to models that are best suited for their type/style of questions. In addition, the reward function could be easily extended to include any factors of importance for the user e.g., GPU memory, environmental impact, licensing or other LLM usage constrains.

%%%%%%%%%%%%%%%%%%%%%%%%%%%%%%%%%%%%%%
%% 9. Conclusion
%%%%%%%%%%%%%%%%%%%%%%%%%%%%%%%%%%%%%%

\section{Conclusion}\label{section:conclusion}

In this work, we underline the challenge of VCF customers to select the best-performing LLMs for their application given certain optimization contexts and resource constraints they may face. To tackle this issue, we propose PickLLM, a reinforcement-learning-based mechanism that examines the performance of the available LLMs while guiding queries to eventually the optimal model in terms of cost per query, overall latency, and response quality. Our approach is easily extendable to cover additional contextual optimization goals, while also being extremely computing-resource efficient in comparison to alternatives. Results indicate the adaptability of the proposed framework towards optimizing user querying sessions in terms of hard metrics with improvements of up to $\approx 50-60\%$ for overall session cost and average latency.

\bibliographystyle{ACM-Reference-Format}
\bibliography{sample-base}

\newpage

\appendix

\section{Prompt templates}\label{appendix:a}

We list the prompt template used for ChatGPT 3.5 Turbo when acting as an LLM judge~\cite{zheng2024judging}:

\begin{mdframed}[backgroundcolor=gray!5, linewidth=1pt, linecolor=black, roundcorner=5pt, innertopmargin=\baselineskip, innerbottommargin=\baselineskip, innerrightmargin=1pt, innerleftmargin=1pt, skipabove=\baselineskip, skipbelow=\baselineskip]

\noindent \textbf{[\texttt{System}]}\\
\texttt{Please act as an impartial judge and evaluate the quality of the response provided by an AI assistant to the user question displayed below. 
We provide also a human generated response to use as guidance for your scoring. Rate the response on a scale of 0 to 1 with three decimal accuracy by strictly returning just one number, for example: "0.345".}

\medskip
\noindent \textbf{[\texttt{User Question}]}\\
\textit{\texttt{question}}

\medskip
\noindent \textbf{[\texttt{AI Response]}}\\
\textit{\texttt{ai\_response}}

\medskip
\noindent \textbf{\texttt{[Human Response:]}}\\
\textit{\texttt{human\_response}}
\end{mdframed}

% \medskip

% \noindent
% As reward model\cite{zheng2024judging}:
% \begin{mdframed}[backgroundcolor=gray!5, linewidth=1pt, linecolor=black, roundcorner=5pt, innertopmargin=\baselineskip, innerbottommargin=\baselineskip, innerrightmargin=1pt, innerleftmargin=1pt, skipabove=\baselineskip, skipbelow=\baselineskip]

% \noindent \textbf{[\texttt{System}]}\\
% \texttt{Please act as a reward model and given a user query provided below, evaluate the quality of the response provided by an AI assistant. Rate the response on a scale of 0 to 1 with three decimal accuracy by strictly returning just one number, for example: "0.345".}

% \medskip
% \noindent \textbf{[\texttt{User Question}]}\\
% \textit{\texttt{question}}

% \medskip
% \noindent \textbf{[\texttt{AI Response]}}\\
% \textit{\texttt{ai\_response}}

% \end{mdframed}

\end{document}